%% file: rotor_drag.tex
\documentclass[letterpaper, 10 pt, conference]{ieeeconf}  

\IEEEoverridecommandlockouts                              

\usepackage{amsmath}
\usepackage{amssymb}
\usepackage{balance}
\usepackage{booktabs}
\usepackage{graphicx}
\usepackage{hyperref}
\usepackage{mathtools}
\usepackage{multirow}
\usepackage{relsize}
\usepackage{soul}
\usepackage{todonotes}

\usepackage[font+=small]{caption}
\usepackage[font+=small]{subcaption}
\usepackage{siunitx}
\usepackage{color}

\newcommand{\bVec}[1]{\mathbf{#1}}
\newcommand{\sVec}[1]{\begin{bmatrix} #1 \end{bmatrix}}
\newcommand{\norm}[1]{\left\lVert#1\right\rVert}
\newcommand{\diag}[1]{\operatorname{diag}\left(#1 \right) }

\newcommand{\vect}[3]{{_{\mathsmaller{\mathrm{#2}}}\mathbf{#1}_{\mathsmaller{\mathrm{#3}}}}} 
\newcommand{\vectss}[4]{{_{\mathsmaller{\mathrm{#2}}}\mathbf{#1}_{\mathsmaller{\mathrm{#3}}}^{\mathsmaller{\mathrm{#4}}}}} 
\newcommand{\vecttrans}[3]{{_{\mathsmaller{\mathrm{#2}}}\mathbf{#1}_{\mathsmaller{\mathrm{#3}}}^{\mathsmaller{\top}}}} 
\newcommand{\vectdot}[3]{{_{\mathsmaller{\mathrm{#2}}}\dot{\mathbf{#1}}_{\mathsmaller{\mathrm{#3}}}}} 

\newcommand{\wfr}[0]{\ensuremath{W}} 
\newcommand{\bfr}[0]{\ensuremath{B}} 
\newcommand{\cfr}[0]{\ensuremath{C}} 

\newcommand{\pos}[0]{\bVec{p}} 
\newcommand{\vel}[0]{\bVec{v}} 
\newcommand{\acc}[0]{\bVec{a}} 
\newcommand{\jerk}[0]{\bVec{j}} 
\newcommand{\snap}[0]{\bVec{s}} 

\newcommand{\ori}[1]{\bVec{R}_{\!\mathsmaller{\mathrm{#1}}}} 
\newcommand{\heading}[0]{\psi} 

\newcommand{\bodyrate}[0]{\omega} 
\newcommand{\bodyrates}[0]{\boldsymbol{\bodyrate}} 

\newcommand{\torqueinputs}[0]{\boldsymbol{\tau}} 
\newcommand{\inertia}[0]{\bVec{J}} 
\newcommand{\gyrotorques}[0]{\vect{\torqueinputs}{}{g}} 

\newcommand{\thrust}[0]{c} 
\newcommand{\horzthrustcoeff}[0]{k_h} 

\newcommand{\gravacc}[0]{\ensuremath{g}} 

\newcommand{\dragcoeff}[1]{\ensuremath{d}_{\!\mathsmaller{\mathrm{#1}}}} 
\newcommand{\dragmat}[0]{\bVec{D}} 
\newcommand{\amat}[0]{\bVec{A}} %
\newcommand{\bmat}[0]{\bVec{B}} %

\newcommand{\uvec}[0]{\bVec{e}} 

\newcommand{\abserr}[0]{\ensuremath{E_a}} 
\newcommand{\poserr}[0]{\ensuremath{E_p}} 

\title{\LARGE \bf Differential Flatness of Quadrotor Dynamics Subject to Rotor Drag for Accurate Tracking of High-Speed Trajectories
}

\author{Matthias Faessler$^{1}$, Antonio Franchi$^{2}$, and Davide Scaramuzza$^{1}$
\thanks{This research was supported by the National Centre of Competence in Research (NCCR) Robotics, the SNSF-ERC Starting Grant, the DARPA FLA program, and the European Union's Horizon 2020 research and innovation program under grant agreement No 644271 AEROARMS.}
\thanks{$^{1}$The two authors are with the Robotics and Perception Group, Dep. of Informatics University of Zurich and Dep. of Neuroinformatics of the University of Zurich and ETH Zurich, Switzerland---\url{http://rpg.ifi.uzh.ch}.}%
\thanks{$^{2}$Antonio Franchi is with LAAS-CNRS, Universit\'e de Toulouse, CNRS, Toulouse, France.}%
}

\begin{document}

\maketitle
\thispagestyle{empty}
\pagestyle{empty}


\begin{abstract}
In this paper, we prove that the dynamical model of a quadrotor subject to linear rotor drag effects is differentially flat in its position and heading.
We use this property to compute feed-forward control terms directly from a reference trajectory to be tracked.
The obtained feed-forward terms are then used in a cascaded, nonlinear feedback control law that enables accurate agile flight with quadrotors.
Compared to state-of-the-art control methods, which treat the rotor drag as an unknown disturbance, our method reduces the trajectory tracking error significantly.
Finally, we present a method based on a gradient-free optimization to identify the rotor drag coefficients, which are required to compute the feed-forward control terms.
The new theoretical results are thoroughly validated trough extensive comparative experiments.
\end{abstract}


\input{chapters/introduction.tex}
\input{chapters/nomenclature.tex}
\input{chapters/model.tex}
\input{chapters/differential_flatness.tex}
\input{chapters/control_law.tex}
\input{chapters/drag_coeff_estimation.tex}
\input{chapters/experiments.tex}
\input{chapters/control_comparison.tex}
\input{chapters/conclusion.tex}





\bibliographystyle{IEEEtran}
\balance
\bibliography{../rpg_bib/all}

\end{document}

%% file: chapters/introduction.tex
\section*{Supplementary Material}

\begin{flushleft} 
Video: \href{https://youtu.be/VIQILwcM5PA}{https://youtu.be/VIQILwcM5PA}
\newline
Code: \href{https://github.com/uzh-rpg/rpg_quadrotor_control}{https://github.com/uzh-rpg/rpg\_quadrotor\_control}
\end{flushleft} 
\section{Introduction} \label{sec:introduction}

\subsection{Motivation} \label{sec:motivation}

For several years, quadrotors have proven to be suitable aerial platforms for performing agile flight maneuvers .
Nevertheless, quadrotors are typically controlled by neglecting aerodynamic effects, such as rotor drag, that only become important for non-hover conditions.
These aerodynamic effects are treated as unknown disturbances, which works well when controlling the quadrotor close to hover conditions but reduces its trajectory tracking accuracy progressively with increasing speed.
For fast obstacle avoidance it is important to perform accurate agile trajectory tracking.
To achieve this, we require a method for accurate tracking of trajectories that are unknown prior to flying.

\begin{figure}[t]
   \centering
   \includegraphics[width=0.5\textwidth]{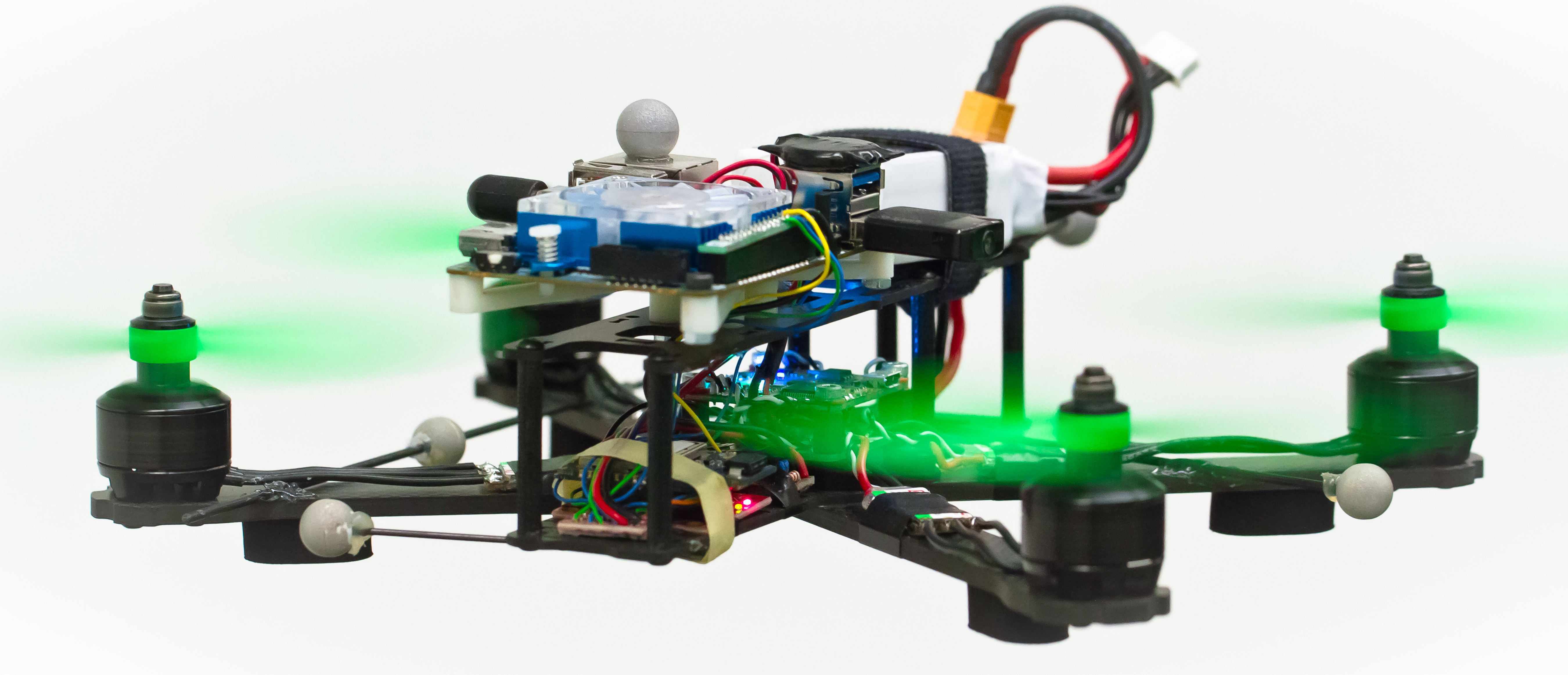}
   \caption{First-person-view racing inspired quadrotor platform used for the presented experiments.}
   \label{fig:quad_picture}
\end{figure}

The main aerodynamic effect causing trajectory tracking errors during high-speed flight is rotor drag, which is a linear effect in the quadrotor's velocity~\cite{Burri17jfr}.
In this work, we aim at developing a control method that improves the trajectory tracking performance of quadrotors by considering the rotor drag effect.
To achieve this, we first prove that the dynamical model of a quadrotor subject to linear rotor drag effects is differentially flat with flat outputs chosen to be its position and heading.
We then use this property to compute feed-forward control terms directly from the reference trajectory to be tracked.
The obtained feed-forward terms are then used in a cascaded, nonlinear feedback control law that enables accurate agile flight with quadrotors on a priori unknown trajectories.
Finally, we present a method based on a gradient-free optimization to identify the rotor drag coefficients which are required to compute the feed-forward control terms.
We validate our theoretical results through experiments with a quadrotor shown in Fig.~\ref{fig:quad_picture}.

\subsection{Related Work} \label{sec:related_work}

In~\cite{Mellinger11icra}, it was shown that the common model of a quadrotor \emph{without} considering rotor drag effects is differentially flat when choosing its position and heading as flat outputs.
Furthermore, this work presented a control algorithm that computes the desired collective thrust and torque inputs from the measured position, velocity, orientation, and body-rates errors.
With this method, agile maneuvers with speeds of several meters per second were achieved.
In~\cite{Ferrin11iros}, the differential flatness property of a hexarotor that takes the desired collective thrust and its desired orientation as inputs was exploited to compute feed-forward terms used in an LQR feedback controller.
The desired orientation was then controlled by a separate low-level control loop, which also enables the execution of flight maneuvers with speeds of several meters per second.
We extend these works by showing that the dynamics of a quadrotor are differentially flat even when they are subject to linear rotor drag effects.
Similarly to~\cite{Ferrin11iros}, we make use of this property to compute feed-forward terms that are then applied by a position controller.

Rotor drag effects influencing a quadrotor's dynamics were investigated in~\cite{Bristeau09ecc} and~\cite{Martin10icra} where also a control law was presented, which considers these dynamics.
Rotor drag effects originate from blade flapping and induced drag of the rotors, which are, thanks to their equivalent mathematical expression, typically combined as linear effects in a lumped parameter dynamical model~\cite{Mahony12ram}.
These rotor drag effects were then incorporated in dynamical models of multi rotors to improve state estimation in~\cite{Leishman14cs} and~\cite{Burri15icra}.
In this work, we make use of the fact that the main aerodynamic effects are of similar nature and can therefore be described together by lumped parameters in a dynamical model.

In~\cite{Bangura17tro}, the authors achieve accurate thrust control by electronic speed controllers through a model of the aerodynamic power generated by a fixed-pitch rotor under wind disturbances, which reduces the trajectory tracking error of a quadrotor.
Rotor drag was also considered in control methods for multi-rotor vehicles in~\cite{Kai17ifac} and~\cite{Omari13iros}, where the control problem was simplified by decomposing the rotor drag force into a component that is independent of the vehicle's orientation and one along the thrust direction, which leads to an explicit expression for the desired thrust direction.
In~\cite{Svacha17icuas}, a refined thrust model and a control scheme that considers rotor drag in the computation of the thrust command and the desired orientation are presented.
Additionally to the thrust command and desired orientation, the control scheme in~\cite{Bangura17thesis} also computes the desired body rates and angular accelerations by considering rotor drag but requires estimates of the quadrotor's acceleration and jerk, which are usually not available.
In contrast, we compute the exact reference thrust, orientation, body rates, and angular accelerations considering rotor drag only from a reference trajectory, which we then use as feed-forward terms in the controller.

%% file: chapters/nomenclature.tex
\section{Nomenclature} \label{sec:nomenclature}

\begin{figure}[t]
   \centering
   \includegraphics[width=0.5\textwidth]{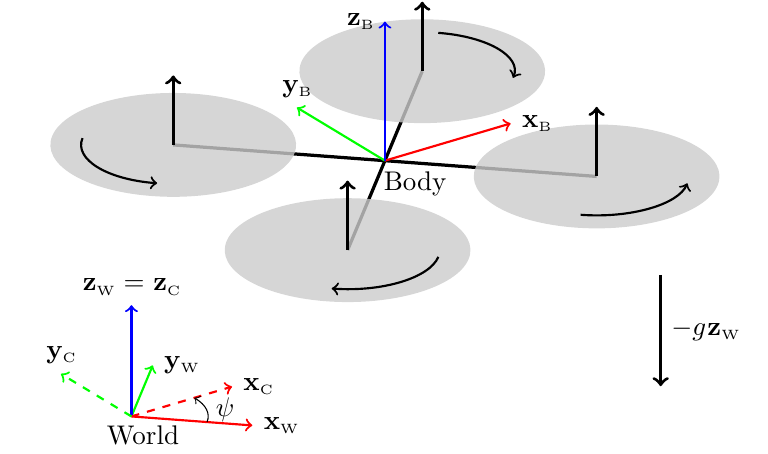}
   \caption{Schematics of the considered quadrotor model with the used coordinate systems.}
   \label{fig:quad_schematics}
\end{figure}

In this work, we make use of a world frame $\wfr$ with orthonormal basis ${\left\{ \vect{x}{}{\wfr}, \vect{y}{}{\wfr}, \vect{z}{}{\wfr} \right\}}$ represented in world coordinates and a body frame $\bfr$ with orthonormal basis ${\left\{ \vect{x}{}{\bfr}, \vect{y}{}{\bfr}, \vect{z}{}{\bfr} \right\}}$ also represented in world coordinates.
The body frame is fixed to the quadrotor with an origin coinciding with its center of mass as depicted in Fig.~\ref{fig:quad_schematics}.
The quadrotor is subject to a gravitational acceleration $\gravacc$ in the $-\vect{z}{}{\wfr}$ direction.
We denote the position of the quadrotor's center of mass as $\pos$, and its derivatives, velocity, acceleration, jerk, and snap as $\vel$, $\acc$, $\jerk$, and $\snap$, respectively.
We represent the quadrotor's orientation as a rotation matrix ${\ori{} = \sVec{\vect{x}{}{\bfr} & \vect{y}{}{\bfr} & \vect{z}{}{\bfr}}}$ and its body rates (i.e., the angular velocity) as $\bodyrates$ represented in body coordinates.
To denote a unit vector along the $z$-coordinate axis we write $\vect{\uvec}{}{z}$.
Finally, we denote quantities that can be computed from a reference trajectory as \emph{reference values} and quantities that are computed by an outer loop feedback control law and passed to an inner loop controller as \emph{desired values}.

%% file: chapters/model.tex
\section{Model} \label{sec:model}

We consider the dynamical model of a quadrotor with rotor drag developed in~\cite{Kai17ifac} with no wind, stiff propellers, and no dependence of the rotor drag on the thrust.
According to this model, the dynamics of the position $\pos$, velocity $\vel$, orientation $\ori{}$, and body rates $\bodyrates$ can be written as
\begin{align}
	\dot{\pos} &= \vel \label{eq:position_dynamics} \\
	\dot{\vel} &= - \gravacc \vect{z}{}{\wfr} + \thrust \vect{z}{}{\bfr} - \ori{} \dragmat \vecttrans{\ori{}}{}{} \vel \label{eq:velocity_dynamics} \\
	\dot{\ori{}} &= \ori{} \hat{\bodyrates} \label{eq:orientation_dynamics} \\
	\dot{\bodyrates} &= \inertia^{-1} \left( \torqueinputs - \bodyrates \times \inertia \bodyrates - \gyrotorques - \amat \vecttrans{\ori{}}{}{} \vel - \bmat \bodyrates \right) \label{eq:body_rate_dynamics}
\end{align}
where $\thrust$ is the mass-normalized collective thrust, ${\dragmat = \diag{\dragcoeff{x}, \dragcoeff{y}, \dragcoeff{z}}}$ is a constant diagonal matrix formed by the mass-normalized rotor-drag coefficients, $\hat{\bodyrates}$ is a skew-symmetric matrix formed from $\bodyrates$, $\inertia$ is the quadrotor's inertia matrix, $\torqueinputs$ is the three dimensional torque input, $\gyrotorques$ are gyroscopic torques from the propellers, and $\amat$ and $\bmat$ are constant matrices.
For the derivations and more details about these terms, please refer to~\cite{Kai17ifac}.
In this work, we adopt the thrust model presented in~\cite{Svacha17icuas}
\begin{equation}
	\thrust = \thrust_{\mathsmaller{\mathrm{cmd}}} + \horzthrustcoeff v_h^2 \label{eq:thrust_model}
\end{equation}
where $\thrust_{\mathsmaller{\mathrm{cmd}}}$ is the commanded collective thrust input, $\horzthrustcoeff$ is a constant, and ${v_h = \vecttrans{v}{}{}( \vect{x}{}{\bfr} + \vect{y}{}{\bfr})}$.
The term $\horzthrustcoeff v_h^2$ acts as a quadratic velocity-dependent input disturbance which adds up to the input $\thrust_{\mathsmaller{\mathrm{cmd}}}$.
The additional linear velocity-dependent disturbance in the $\vect{z}{}{\bfr}$ direction of the thrust model in~\cite{Svacha17icuas} is lumped by $\dragcoeff{z}$ directly in~\eqref{eq:velocity_dynamics} by neglecting its dependency on the rotor speeds.
Note that this dynamical model of a quadrotor is a generalization of the common model found, e.g., in~\cite{Mellinger11icra}, in which the linear rotor drag components are typically neglected, i.e., $\dragmat$, $\amat$ and $\bmat$ are considered null matrices.

%% file: chapters/differential_flatness.tex
\section{Differential Flatness} \label{sec:differential_flatness}

In this section, we show that the extended dynamical model of a quadrotor subject to rotor drag~\eqref{eq:position_dynamics}-\eqref{eq:body_rate_dynamics} with four inputs is differentially flat, like the model with neglected drag~\cite{Mellinger11icra}.
In fact, we shall show that the states $[\pos, \vel, \ori{}, \bodyrates]$ and the inputs $[\thrust_{\mathsmaller{\mathrm{cmd}}}, \torqueinputs]$ 
can be written as algebraic functions of four selected flat outputs and a finite number of their derivatives.
Equally to~\cite{Mellinger11icra}, we choose the flat outputs to be the quadrotor's position $\pos$ and its heading $\heading$.

To show that the orientation $\ori{}$ and the collective thrust $\thrust$ are functions of the flat outputs, we reformulate~\eqref{eq:velocity_dynamics} as
\begin{align}
	&\thrust \vect{z}{}{\bfr} - 
	\left( \dragcoeff{x} \vecttrans{x}{}{\bfr} \vel \right) \vect{x}{}{\bfr} -
	\left( \dragcoeff{y} \vecttrans{y}{}{\bfr} \vel \right) \vect{y}{}{\bfr} -
	\left( \dragcoeff{z} \vecttrans{z}{}{\bfr} \vel \right) \vect{z}{}{\bfr} \nonumber \\
	& \qquad - \acc - \gravacc \vect{z}{}{\wfr} = 0 . \label{eq:vel_dyn_reorg}
\end{align}
From left-multiplying~\eqref{eq:vel_dyn_reorg} by $\vecttrans{x}{}{\bfr}$ we get
\begin{equation}
	\vecttrans{x}{}{\bfr} \boldsymbol{\alpha} = 0, \; \text{with} \quad \boldsymbol{\alpha} = \acc + \gravacc \vect{z}{}{\wfr} + \dragcoeff{x} \vel . \label{eq:x_body_constraint}
\end{equation}
From left-multiplying~\eqref{eq:vel_dyn_reorg} by $\vecttrans{y}{}{\bfr}$ we get
\begin{equation}
	\vecttrans{y}{}{\bfr} \boldsymbol{\beta} = 0, \; \text{with} \quad \boldsymbol{\beta} = \acc + \gravacc \vect{z}{}{\wfr} + \dragcoeff{y} \vel . \label{eq:y_body_constraint}
\end{equation}
To enforce a reference heading $\heading$, we constrain the projection of the $\vect{x}{}{\bfr}$ axis into the $\vect{x}{}{\wfr} - \vect{y}{}{\wfr}$ plane to be collinear with $\vect{x}{}{\cfr}$ (cf. Fig.~\ref{fig:quad_schematics}), where
\begin{align}
	\vect{x}{}{\cfr} &= \begin{bmatrix}
		\cos(\heading) & \sin(\heading) & 0
	\end{bmatrix}^{\top} \\
	\vect{y}{}{\cfr} &= \begin{bmatrix}
		-\sin(\heading) & \cos(\heading) & 0
	\end{bmatrix}^{\top} . \label{eq:y_c}
\end{align}
From this, \eqref{eq:x_body_constraint} and \eqref{eq:y_body_constraint}, and the constraints that $\vect{x}{}{\bfr}$, $\vect{y}{}{\bfr}$, and $\vect{z}{}{\bfr}$ must be orthogonal to each other and of unit length, we can construct $\ori{}$ with
\begin{align}
	\vect{x}{}{\bfr} &= \frac{\vect{y}{}{\cfr} \times \boldsymbol{\alpha}}{\norm{\vect{y}{}{\cfr} \times \boldsymbol{\alpha}}} \\
	\vect{y}{}{\bfr} &= \frac{\boldsymbol{\beta} \times \vect{x}{}{\bfr}}{\norm{\boldsymbol{\beta} \times \vect{x}{}{\bfr}}} \\
	\vect{z}{}{\bfr} &= \vect{x}{}{\bfr} \times \vect{y}{}{\bfr} .
\end{align}
One can verify that these vectors are of unit length, perpendicular to each other, and satisfy the constraints~\eqref{eq:x_body_constraint} - \eqref{eq:y_c}.
To get the collective thrust, we left-multiply~\eqref{eq:vel_dyn_reorg} by $\vecttrans{z}{}{\bfr}$
\begin{equation}
	\thrust = \vecttrans{z}{}{\bfr} \left( \acc + \gravacc \vect{z}{}{\wfr} + \dragcoeff{z} \vel \right) .
\end{equation}
Then the collective thrust input can be computed as a function of $\thrust$, $\ori{}$, and the flat outputs, as
\begin{equation}
	\thrust_{\mathsmaller{\mathrm{cmd}}} = \thrust - \horzthrustcoeff (\vecttrans{v}{}{}( \vect{x}{}{\bfr} + \vect{y}{}{\bfr}))^2 .
\end{equation}
To show that the body rates $\bodyrates$ are functions of the flat outputs and their derivatives, we take the derivative of~\eqref{eq:velocity_dynamics}
\begin{equation}
	\jerk = \dot{\thrust} \vect{z}{}{\bfr} + \thrust \ori{} \hat{\bodyrates} \vect{\uvec}{}{z} - \ori{} \left( \left( \hat{\bodyrates} \dragmat + \dragmat \vecttrans{\hat{\bodyrates}}{}{} \right) \vecttrans{\ori{}}{}{} \vel + \dragmat \vecttrans{\ori{}}{}{} \acc \right) . \label{eq:vel_dyn_diff}
\end{equation}
Left-multiplying~\eqref{eq:vel_dyn_diff} by $\vecttrans{x}{}{\bfr}$ and rearranging terms, we get
\begin{align}
	&\bodyrate_y \left( \thrust - \left( \dragcoeff{z} - \dragcoeff{x} \right) \left( \vecttrans{z}{}{\bfr} \vel \right) \right) - \bodyrate_z \left( \dragcoeff{x} - \dragcoeff{y} \right) \left( \vecttrans{y}{}{\bfr} \vel \right) \nonumber \\
	& \quad = \vecttrans{x}{}{\bfr} \jerk + \dragcoeff{x} \vecttrans{x}{}{\bfr} \acc . \label{eq:pitch_rate_cond}
\end{align}
Left-multiplying~\eqref{eq:vel_dyn_diff} by $\vecttrans{y}{}{\bfr}$ and rearranging terms, we get
\begin{align}
	&\bodyrate_x \left( \thrust + \left( \dragcoeff{y} - \dragcoeff{z} \right) \left( \vecttrans{z}{}{\bfr} \vel \right) \right) + \bodyrate_z \left( \dragcoeff{x} - \dragcoeff{y} \right) \left( \vecttrans{x}{}{\bfr} \vel \right) \nonumber \\
	& \quad = -\vecttrans{y}{}{\bfr} \jerk - \dragcoeff{y} \vecttrans{y}{}{\bfr} \acc . \label{eq:roll_rate_cond}
\end{align}
To get a third constraint for the body rates, we project~\eqref{eq:orientation_dynamics} along $\vect{y}{}{\bfr}$
\begin{equation}
	\bodyrate_z = \vecttrans{y}{}{\bfr} \vectdot{x}{}{\bfr} . \label{eq:yaw_rate_exp}
\end{equation}
Since $\vect{x}{}{\bfr}$ is perpendicular to $\vect{y}{}{\cfr}$ and $\vect{z}{}{\bfr}$, we can write
\begin{equation}
	\vect{x}{}{\bfr} = \frac{\vect{\tilde{x}}{}{\bfr}}{\norm{\vect{\tilde{x}}{}{\bfr}}}, \; \text{with} \quad \vect{\tilde{x}}{}{\bfr} = \vect{y}{}{\cfr} \times \vect{z}{}{\bfr} . \label{eq:x_body_tilde_def}
\end{equation}
Taking its derivative as the general derivative of a normalized vector, we get
\begin{equation}
	\vectdot{x}{}{\bfr} = \frac{\vect{\dot{\tilde{x}}}{}{\bfr}}{\norm{\vect{\tilde{x}}{}{\bfr}}} - \vect{\tilde{x}}{}{\bfr} \frac{\vecttrans{\tilde{x}}{}{\bfr} \vect{\dot{\tilde{x}}}{}{\bfr}}{\norm{\vect{\tilde{x}}{}{\bfr}}^3}
\end{equation}
and, since $\vect{\tilde{x}}{}{\bfr}$ is collinear to $\vect{x}{}{\bfr}$ and therefore perpendicular to $\vect{y}{}{\bfr}$, we can write~\eqref{eq:yaw_rate_exp} as
\begin{equation}
	\bodyrate_z = \vecttrans{y}{}{\bfr} \frac{\vect{\dot{\tilde{x}}}{}{\bfr}}{\norm{\vect{\tilde{x}}{}{\bfr}}} . \label{eq:yaw_rate_subs}
\end{equation}
The derivative of $\vect{\tilde{x}}{}{\bfr}$ can be computed as
\begin{align}
	\vect{\dot{\tilde{x}}}{}{\bfr} &= \vect{\dot{y}}{}{\cfr} \times \vect{z}{}{\bfr} + \vect{y}{}{\cfr} \times \vect{\dot{z}}{}{\bfr} , \\
	&= \left( -\dot{\heading} \vect{x}{}{\cfr} \right) \times \vect{z}{}{\bfr} + \vect{y}{}{\cfr} \times \left( \bodyrate_y \vect{x}{}{\bfr} - \bodyrate_x \vect{y}{}{\bfr} \right) .
\end{align}
From this, \eqref{eq:x_body_tilde_def}, \eqref{eq:yaw_rate_subs}, and the vector triple product $\vecttrans{a}{}{} (\bVec{b} \times \bVec{c}) = -\vecttrans{b}{}{} (\bVec{a} \times \bVec{c})$ we then get
\begin{equation}
	\bodyrate_z = \frac{1}{\norm{\vect{y}{}{\cfr} \times \vect{z}{}{\bfr}}} \left( \dot{\heading} \vecttrans{x}{}{\cfr} \vect{x}{}{\bfr} + \bodyrate_y \vecttrans{y}{}{\cfr} \vect{z}{}{\bfr} \right) . \label{eq:yaw_rate_cond}
\end{equation}
The body rates can now be obtained by solving the linear system of equations composed of~\eqref{eq:pitch_rate_cond}, \eqref{eq:roll_rate_cond}, and~\eqref{eq:yaw_rate_cond}.

To compute the angular accelerations $\dot{\bodyrates}$ as functions of the flat outputs and their derivatives, we take the derivative of~\eqref{eq:pitch_rate_cond}, \eqref{eq:roll_rate_cond}, and~\eqref{eq:yaw_rate_cond} to get a similar linear system of equations as
\begin{align}
	&\dot{\bodyrate}_y \left( \thrust - \left( \dragcoeff{z} - \dragcoeff{x} \right) \left( \vecttrans{z}{}{\bfr} \vel \right) \right) - \dot{\bodyrate}_z \left( \dragcoeff{x} - \dragcoeff{y} \right) \left( \vecttrans{y}{}{\bfr} \vel \right) \nonumber \\
	&\quad = \vecttrans{x}{}{\bfr} \snap - 2 \dot{\thrust} \bodyrate_y - \thrust \bodyrate_x \bodyrate_z + \vecttrans{x}{}{\bfr} \boldsymbol{\xi} \\
	&\dot{\bodyrate}_x \left( \thrust + \left( \dragcoeff{y} - \dragcoeff{z} \right) \left( \vecttrans{z}{}{\bfr} \vel \right) \right) + \dot{\bodyrate}_z \left( \dragcoeff{x} - \dragcoeff{y} \right) \left( \vecttrans{x}{}{\bfr} \vel \right) \nonumber \\
	&\quad = - \vecttrans{y}{}{\bfr} \snap - 2 \dot{\thrust} \bodyrate_x + \thrust \bodyrate_y \bodyrate_z - \vecttrans{y}{}{\bfr} \boldsymbol{\xi} \\
	- &\dot{\bodyrate}_y \vecttrans{y}{}{\cfr} \vect{z}{}{\bfr} + \dot{\bodyrate}_z \norm{\vect{y}{}{\cfr} \times \vect{z}{}{\bfr}} \nonumber \\
	&\quad = \ddot{\heading} \vecttrans{x}{}{\cfr} \vect{x}{}{\bfr} + 2 \dot{\heading} \bodyrate_z \vecttrans{x}{}{\cfr} \vect{y}{}{\bfr} - 2 \dot{\heading} \omega_y \vecttrans{x}{}{\cfr} \vect{z}{}{\bfr} \nonumber \\
	& \qquad - \bodyrate_x \bodyrate_y \vecttrans{y}{}{\cfr} \vect{y}{}{\bfr} - \bodyrate_x \bodyrate_z \vecttrans{y}{}{\cfr} \vect{z}{}{\bfr}
\end{align}
which we can solve for $\dot{\bodyrates}$ with
\begin{align}
	\dot{\thrust} &= \vecttrans{z}{}{\bfr} \jerk + \bodyrate_x \left( \dragcoeff{y} - \dragcoeff{z} \right) \left( \vecttrans{y}{}{\bfr} \vel \right) \nonumber \\
	& \quad + \bodyrate_y \left( \dragcoeff{z} - \dragcoeff{x} \right) \left( \vecttrans{x}{}{\bfr} \vel \right) + \dragcoeff{z} \vecttrans{z}{}{\bfr} \acc \\
	\boldsymbol{\xi} &= \ori{} \left( \vectss{\hat{\bodyrates}}{}{}{2} \dragmat + \dragmat \vectss{\hat{\bodyrates}}{}{}{2} + 2 \hat{\bodyrates} \dragmat \vecttrans{\hat{\bodyrates}}{}{} \right) \vecttrans{\ori{}}{}{} \vect{\vel}{}{} \nonumber \\
	& \quad + 2 \ori{} \left( \hat{\bodyrates} \dragmat + \dragmat \vecttrans{\hat{\bodyrates}}{}{} \right) \vecttrans{\ori{}}{}{} \vect{\acc}{}{} + \ori{} \dragmat \vecttrans{\ori{}}{}{} \vect{\jerk}{}{} .
\end{align}
Once we know the angular accelerations, we can solve~\eqref{eq:body_rate_dynamics} for the torque inputs $\torqueinputs$.

Note that, besides quadrotors, this proof also applies to multi-rotor vehicles with parallel rotor axes in general.
More details of this proof can be found in our technical report~\cite{Faessler17tr}.

%% file: chapters/control_law.tex
\section{Control Law} \label{sec:control_law}

To track a reference trajectory, we use a controller consisting of feedback terms computed from tracking errors as well as feed-forward terms computed from the reference trajectory using the quadrotor's differential flatness property.
Apart from special cases, the control architectures of typical quadrotors do not allow to apply the torque inputs directly.
They instead provide a low-level body-rate controller, which accepts desired body rates.
In order to account for this possibility, we designed our control algorithm with a classical cascaded structure, i.e., consisting of a high-level position controller and the low-level body-rate controller.
The high-level position controller computes the desired orientation $\ori{des}$, the collective thrust input $\thrust_{\mathsmaller{\mathrm{cmd}}}$, the desired body rates $\vect{\bodyrates}{}{des}$, and the desired angular accelerations $\vectdot{\bodyrates}{}{des}$, which are then applied in a low-level controller (e.g. as presented in~\cite{Faessler17ral}).
As a first step in the position controller, we compute the desired acceleration of the quadrotor's body as
\begin{align}
	\vect{\acc}{}{des} &= 
	\vect{\acc}{}{fb}
	+ \vect{\acc}{}{ref}
	- \vect{\acc}{}{rd}
	+ \gravacc \vect{z}{}{\wfr}
	\label{eq:position_controller}
\end{align}
where $\vect{\acc}{}{fb}$ are the PD feedback-control terms computed from the position and velocity control errors as 
\begin{equation}
	\vect{\acc}{}{fb} = -\vect{K}{}{pos} \left( \pos - \vect{\pos}{}{ref} \right) - \vect{K}{}{vel} \left( \vel - \vect{\vel}{}{ref} \right)
\end{equation}
where $\vect{K}{}{pos}$ and $\vect{K}{}{vel}$ are constant diagonal matrices and ${\vect{\acc}{}{rd} = - \ori{ref} \dragmat \vecttrans{\ori{}}{}{ref} \vect{\vel}{}{ref}}$ are the accelerations due to rotor drag.
We compute the desired orientation $\ori{des}$ such that the desired acceleration $\vect{\acc}{}{des}$ and the reference heading $\vect{\heading}{}{ref}$ is respected as 
\begin{align}
	\vect{z}{}{\bfr,des} &= \frac{\vect{\acc}{}{des}}{\norm{\vect{\acc}{}{des}}} \\
	\vect{x}{}{\bfr,des} &= \frac{\vect{y}{}{\cfr} \times \vect{z}{}{\bfr,des}}{\norm{\vect{y}{}{\cfr} \times \vect{z}{}{\bfr,des}}} \\
	\vect{y}{}{\bfr,des} &= \vect{z}{}{\bfr,des} \times \vect{x}{}{\bfr,des} .
\end{align}
By projecting the desired accelerations onto the actual body $z$-axis and considering the thrust model~\eqref{eq:thrust_model}, we can then compute the collective thrust input as
\begin{equation}
	\thrust_{\mathsmaller{\mathrm{cmd}}} = \vecttrans{\acc}{}{des} \vect{z}{}{\bfr} - \horzthrustcoeff (\vecttrans{v}{}{}( \vect{x}{}{\bfr}+ \vect{y}{}{\bfr}))^2.
	\label{eq:norm_thrust}
\end{equation}
Similarly, we can compute the desired body rates as
\begin{equation}
	\vect{\bodyrates}{}{des} = \vect{\bodyrates}{}{fb} + \vect{\bodyrates}{}{ref}
\end{equation}
where $\vect{\bodyrates}{}{fb}$ are the feedback terms computed from an attitude controller (e.g. as presented in~\cite{Faessler15icra}) and $\vect{\bodyrates}{}{ref}$ are feed-forward terms from the reference trajectory, which are computed as described in Section~\ref{sec:differential_flatness}.
Finally, the desired angular accelerations are the reference angular accelerations
\begin{equation}
	\vect{\dot{\bodyrates}}{}{des} = \vect{\dot{\bodyrates}}{}{ref}
\end{equation}
which are computed from the reference trajectory as described in Section~\ref{sec:differential_flatness}.

%% file: chapters/drag_coeff_estimation.tex
\section{Drag Coefficients Estimation} \label{sec:drag_coeff_estimation}

To apply the presented control law with inputs $[\thrust_{\mathsmaller{\mathrm{cmd}}}, \bodyrates]$, we need to identify $\dragmat$, and $\horzthrustcoeff$, which are used to compute the reference inputs and the thrust command.
If instead one is using a platform that is controlled by the inputs $[\thrust_{\mathsmaller{\mathrm{cmd}}}, \torqueinputs]$, $\amat$ and $\bmat$ also need to be identified for computing the torque input.
While $\dragmat$ and $\horzthrustcoeff$ can be accurately estimated from measured accelerations and velocities through~\eqref{eq:velocity_dynamics} and~\eqref{eq:thrust_model}, the effects of $\amat$ and $\bmat$ on the body-rate dynamics~\eqref{eq:body_rate_dynamics} are weaker and require to differentiate the gyro measurements as well as knowing the rotor speeds and rotor inertia to compute $\torqueinputs$ and $\gyrotorques$.
Therefore, we propose to identify $\dragmat$, $\amat$, $\bmat$, and $\horzthrustcoeff$ by running a Nelder-Mead gradient free optimization~\cite{Nelder65comjnl} for which the quadrotor repeats a predefined trajectory in each iteration of the optimization.
During this procedure, we control the quadrotor by the proposed control scheme with different drag coefficients in each iteration during which we record the absolute trajectory tracking error~\eqref{eq:absolute_error} and use it as cost for the optimization.
Once the optimization has converged, we know the coefficients that reduce the trajectory tracking error the most.
We found that the obtained values for $\dragmat$ agree with an estimation through~\eqref{eq:velocity_dynamics} when recording IMU measurements and ground truth velocity.
This procedure has the advantage that no IMU and rotor speed measurements are required, which are both unavailable on our quadrotor platform used for the presented experiments, and the gyro measurements do not need to be differentiated.
This is not the case when performing the identification through~\eqref{eq:velocity_dynamics}, \eqref{eq:body_rate_dynamics}, and \eqref{eq:thrust_model}.
Also, since our method does not rely explicitly on~\eqref{eq:velocity_dynamics}, it can also capture first order approximations of non modeled effects lumped into the identified coefficients.
Furthermore, our implementation allows stopping and restarting the optimization at any time, which allows changing the battery.

%% file: chapters/experiments.tex
\section{Experiments} \label{sec:experiments}

\subsection{Experimental Setup} \label{sec:experimental_setup}

Our quadrotor platform is built from off-the-shelf components used for first-person-view racing (see Fig.~\ref{fig:quad_picture}).
It features a carbon frame with stiff six inch propellers, a Raceflight Revolt flight controller, an Odroid XU4 single board computer, and a Laird RM024 radio module for receiving control commands.
The platform weights \SI{610}{\gram} and has a thrust-to-weight ratio of $4$.
To improve its trajectory tracking accuracy we compensate the thrust commands for the varying battery voltage.
All the presented flight experiments were conducted in an OptiTrack motion capture system to acquire the ground truth state of the quadrotor which is obtained at \SI{200}{\hertz} and is used for control and evaluation of the trajectory tracking performance.
Note that our control method and the rotor-drag coefficient identification also work with state estimates that are obtained differently than with a motion capture system.
We compensate for an average latency of the perception and control pipeline of \SI{32}{\milli\second}.
The high-level control runs on a laptop computer at \SI{55}{\hertz}, sending collective thrust and body rate commands to the on-board flight controller where they are tracked by a PD controller running at \SI{4}{\kilo\hertz}.

To evaluate the trajectory tracking performance and as cost for estimating the drag coefficients, we use the absolute trajectory tracking error defined as the root mean square position error
\begin{equation}
	\abserr = \sqrt{\frac{1}{N} \sum_{k=1}^N \norm{\poserr^k}^2} , \; \text{where} \quad \poserr^k = \vectss{\pos}{}{}{k} - \vectss{\pos}{}{ref}{k} \label{eq:absolute_error}
\end{equation}
over $N$ control cycles of the high-level controller required to execute a given trajectory.

\subsection{Trajectories} \label{sec:trajectories}

For identifying the rotor-drag coefficients and for demonstrating the trajectory tracking performance of the proposed controller, we let the quadrotor execute a horizontal circle trajectory and a horizontal Gerono lemniscate trajectory.
The circle trajectory has a radius of \SI{1.8}{\meter} with a velocity of \SI{4}{\meter\per\second} resulting, for the case of not considering rotor drag, in a required collective mass-normalized thrust of ${\thrust = \SI{13.24}{\meter\per\second\squared}}$ and a maximum body rates norm of ${\norm{\bodyrates} = \SI{85}{\degree\per\second}}$.
Its maximum nominal velocity in the $\vect{x}{}{\bfr}$ and $\vect{y}{}{\bfr}$ is \SI{4.0}{\meter\per\second} and \SI{0.0}{\meter\per\second} in the $\vect{z}{}{\bfr}$ direction.
The Gerono lemniscate trajectory is defined by $\left[ {x(t) = 2 \cos \left( \sqrt{2} t \right)}; {y(t) = 2 \sin \left( \sqrt{2} t \right) \cos \left( \sqrt{2} t \right)} \right]$ with a maximum velocity of \SI{4}{\meter\per\second}, a maximum collective mass-normalized thrust of ${\thrust = \SI{12.98}{\meter\per\second\squared}}$, and a maximum body rates norm of ${\norm{\bodyrates} = \SI{136}{\degree\per\second}}$.
Its maximum nominal velocity in the $\vect{x}{}{\bfr}$ and $\vect{y}{}{\bfr}$ is \SI{2.8}{\meter\per\second} and \SI{1.3}{\meter\per\second} in the $\vect{z}{}{\bfr}$ direction for the case of not considering rotor drag.

\subsection{Drag Coefficients Identification} \label{sec:drag_coeff_id}

To identify the drag coefficients in $\dragmat$, we ran the optimization presented in Section~\ref{sec:drag_coeff_estimation} multiple times on both the circle and lemniscate trajectories until it converged, i.e., until the changes of each coefficient in one iteration is below a specified threshold.
We do not identify $\amat$ and $\bmat$ since the quadrotor platform used for the presented experiments takes $[\thrust_{\mathsmaller{\mathrm{cmd}}}, \bodyrates]$ as inputs and we therefore do not need to compute the torque inputs involving $\amat$ and $\bmat$ according to~\eqref{eq:body_rate_dynamics}.
Since we found that $\dragcoeff{z}$ and $\horzthrustcoeff$ only have minor effects on the trajectory tracking performance, we isolate the effects of $\dragcoeff{x}$ and $\dragcoeff{y}$ by setting $\dragcoeff{z} = 0$ and $\horzthrustcoeff = 0$ for the presented experiments.
For each iteration of the optimization, the quadrotor flies two loops of either trajectory.
The optimization typically converges after about 70 iterations, which take around \SI{30}{\minute} including multiple battery swaps.

\begin{figure}[t]
   \centering
   \includegraphics[width=0.48\textwidth]{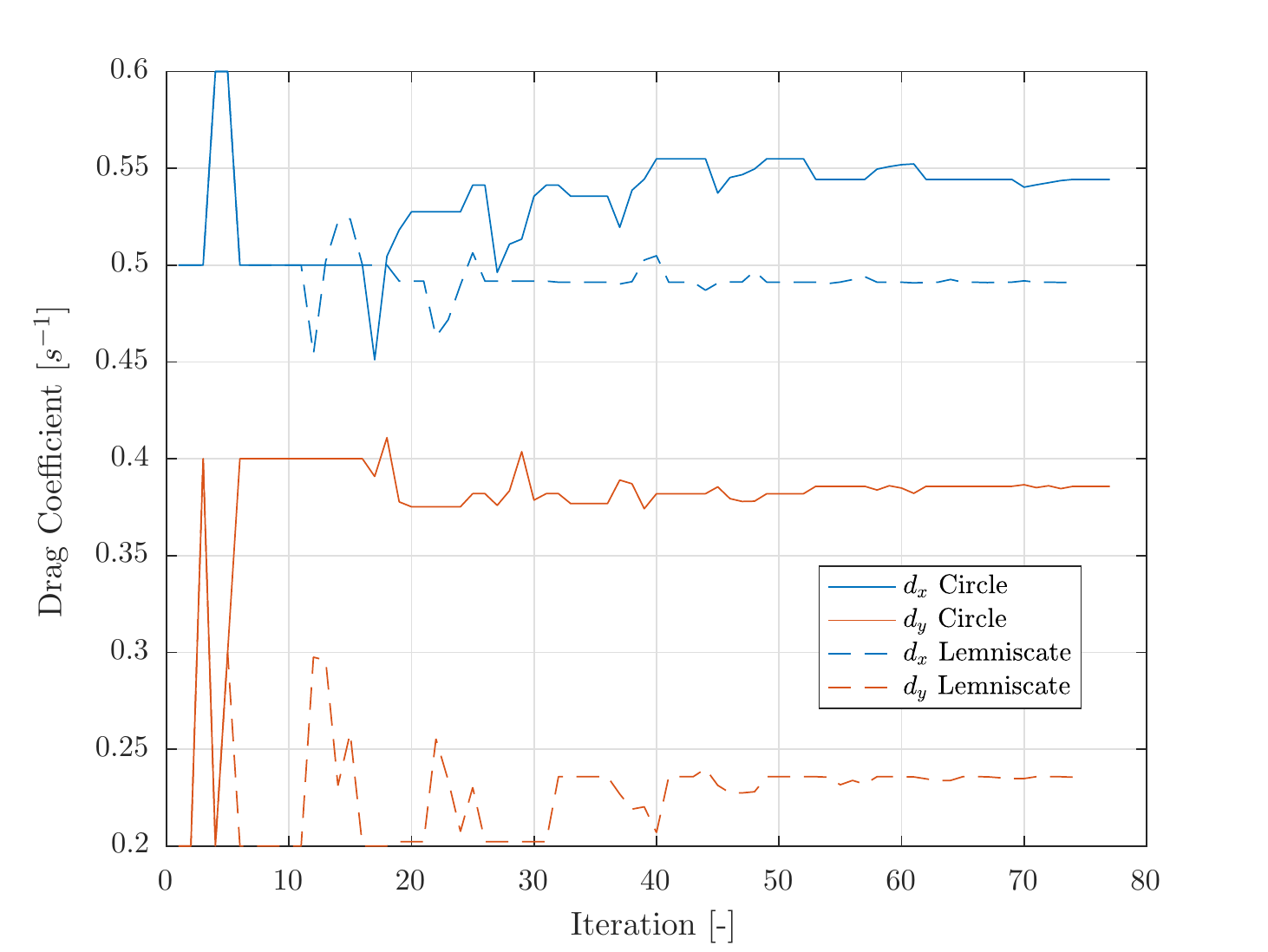}
   \caption{The best performing drag coefficients $\dragcoeff{x}$ and $\dragcoeff{y}$ for every iteration of the identification on both the circle and the lemniscate trajectory.}
   \label{fig:drag_coeff_identification}
\end{figure}

The evolution of the best performing drag coefficients is shown in Fig.~\ref{fig:drag_coeff_identification} for every iteration of the proposed optimization.
On the circle trajectory, we obtained ${\dragcoeff{x} = \SI{0.544}{\per\second}}$ and ${\dragcoeff{y} = \SI{0.386}{\per\second}}$, whereas on the lemniscate trajectory we obtained ${\dragcoeff{x} = \SI{0.491}{\per\second}}$ and ${\dragcoeff{y} = \SI{0.236}{\per\second}}$.
For both trajectories, $\dragcoeff{x}$ is larger than $\dragcoeff{y}$, which is expected because we use a quadrotor that is wider than long.
The obtained drag coefficients identified on the circle are different than the ones identified on the lemniscate, which is due to the fact that the circle trajectory excites velocities in the $\vect{x}{}{\bfr}$ and $\vect{y}{}{\bfr}$ more than the lemniscate trajectory.
We could verify this claim by running the identification on the circle trajectory with a speed of \SI{2.8}{\meter\per\second}, which corresponds to the maximum speeds reached in $\vect{x}{}{\bfr}$ and $\vect{y}{}{\bfr}$ on the lemniscate trajectory.
For this speed, we obtained ${\dragcoeff{x} = \SI{0.425}{\per\second}}$, and ${\dragcoeff{y} = \SI{0.256}{\per\second}}$ on the circle trajectory, which are close to the coefficients identified on the lemniscate trajectory.
Additionally, in our dynamical model, we assume the rotor drag to be independent of the thrust.
This is not true in reality and therefore leads to different results of the drag coefficient estimation on different trajectories where different thrusts are applied.
These reasons suggest to carefully select a trajectory for the identification, which goes towards the problem of finding the optimal trajectory for parameter estimation, which is outside the scope of this paper.
In all the conducted experiments, we found that a non zero drag coefficient in the $z$-direction $\dragcoeff{z}$ does not improve the trajectory tracking performance.
We found this to be true even for purely vertical trajectories with velocities of up to \SI{2.5}{\meter\per\second}.
Furthermore, we found that an estimated $\horzthrustcoeff = \SI{0.009}{\per\meter}$ improves the trajectory tracking performance further but by about one order of magnitude less than $\dragcoeff{x}$ and $\dragcoeff{y}$ on the considered trajectories.

\subsection{Trajectory Tracking Performance} \label{sec:traj_tracking_performance}

\begin{figure}[t]
   \centering
   \includegraphics[width=0.48\textwidth]{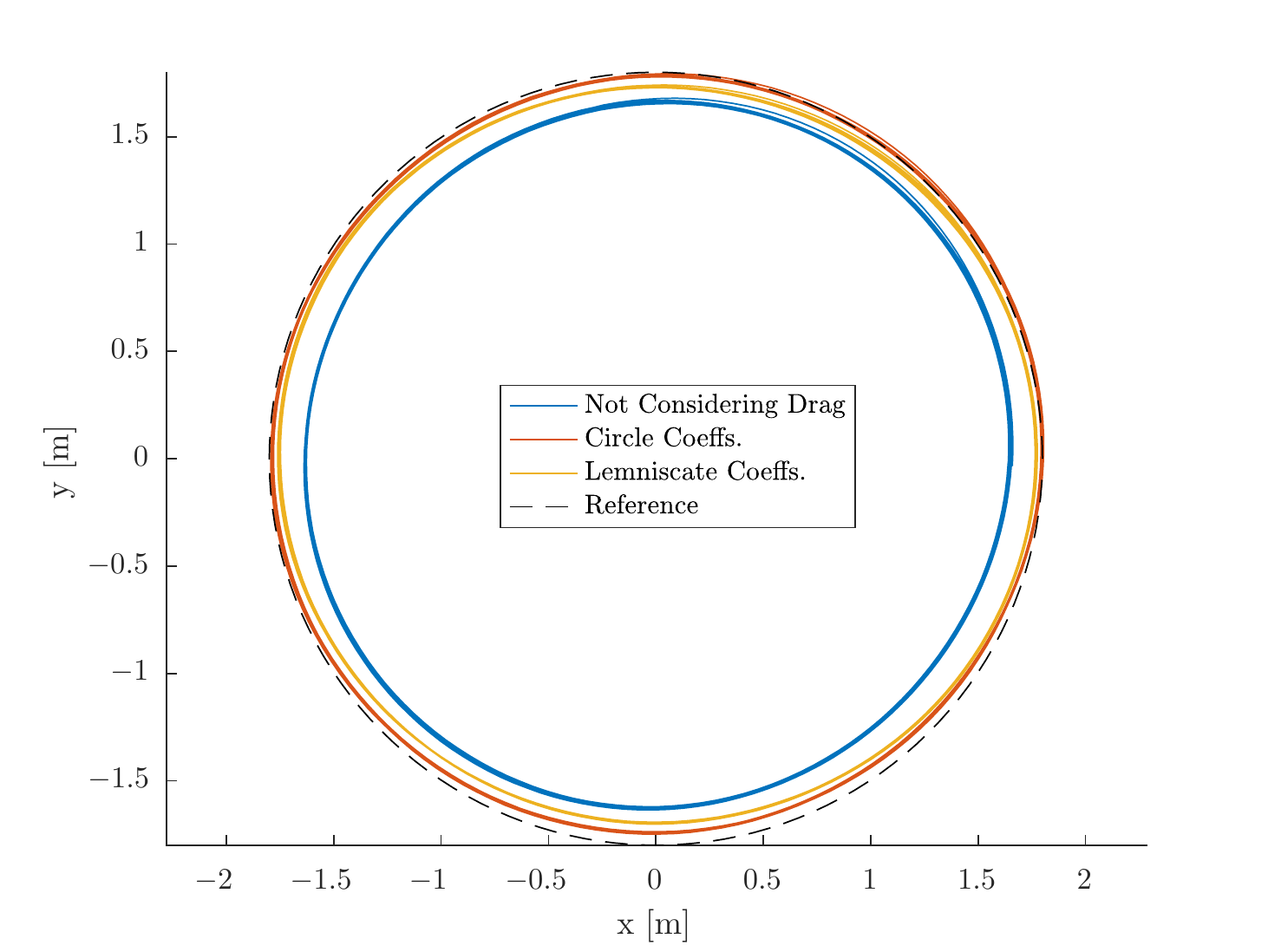}
   \caption{Ground truth position for ten loops on the circle trajectory without considering rotor drag (solid blue), with drag coefficients estimated on the circle trajectory (solid red), and with drag coefficients estimated on the lemniscate trajectory (solid yellow) compared to the reference position (dashed black).}
   \label{fig:circle_traj_performance}
\end{figure}

\begin{figure}[t]
   \centering
   \includegraphics[width=0.48\textwidth]{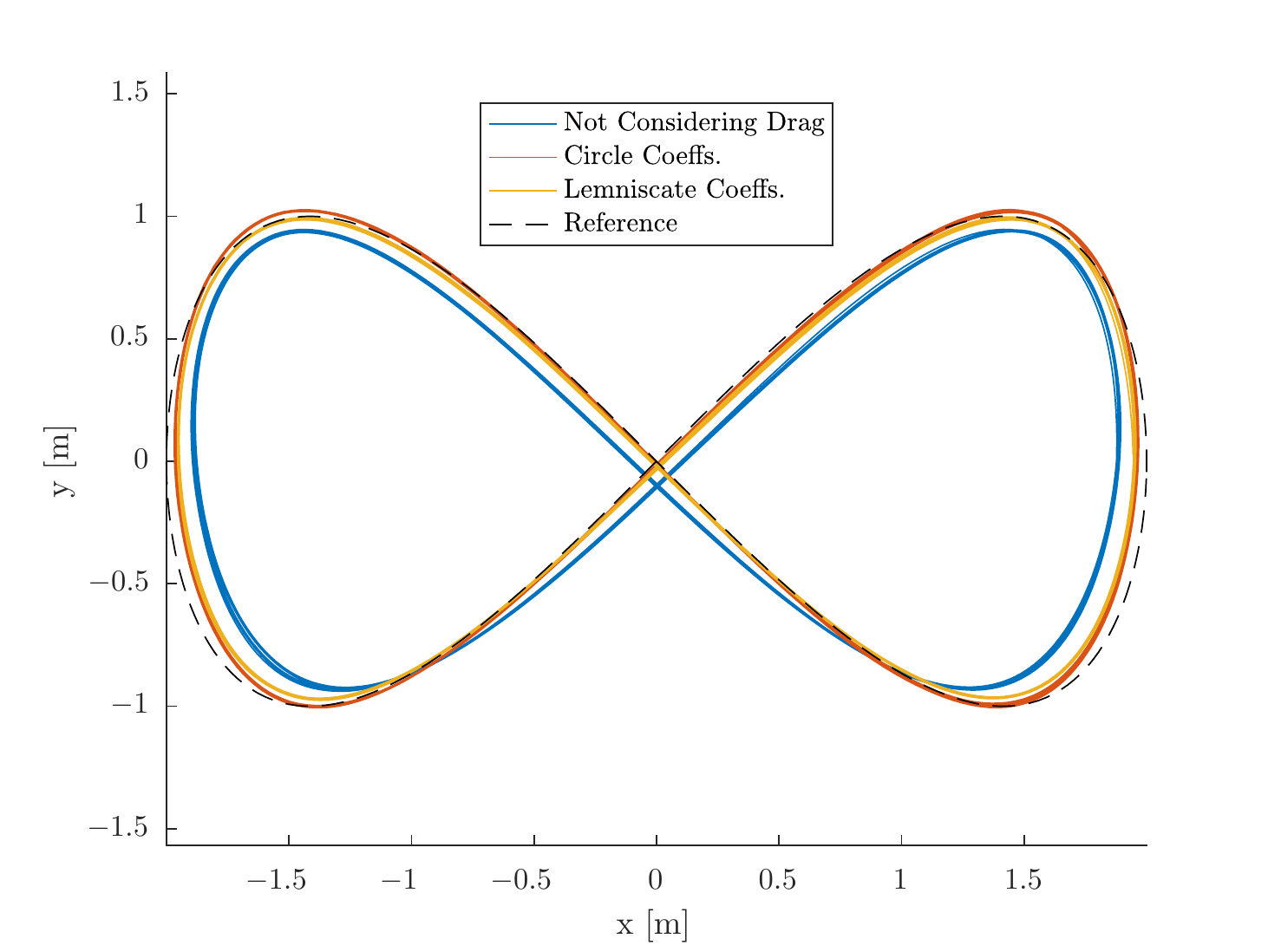}
   \caption{Ground truth position for ten loops on the lemniscate trajectory without considering rotor drag (solid blue), with drag coefficients estimated on the circle trajectory (solid red), and with drag coefficients estimated on the lemniscate trajectory (solid yellow) compared to the reference position (dashed black).}
   \label{fig:figure_eight_traj_performance}
\end{figure}

To demonstrate the trajectory tracking performance of the proposed control scheme, we compare the position error of our quadrotor flying the circle and the lemniscate trajectory described above for three conditions: (i) without considering rotor drag, (ii) with the drag coefficients estimated on the circle trajectory, and (iii) with the drag coefficients estimated on the lemniscate trajectory.\footnote{Video of the experiments: \url{https://youtu.be/VIQILwcM5PA}}
Fig.~\ref{fig:circle_traj_performance} shows the ground truth and reference position when flying the circle trajectory under these three conditions.
Equally, Fig.~\ref{fig:figure_eight_traj_performance} shows the ground truth and reference position when flying the lemniscate trajectory under the same three conditions.
The tracking performance statistics for both trajectories are summarized in Table~\ref{table:trajectory_tracking_statistics}.
From these statistics, we see that the trajectory tracking performance has improved significantly when considering rotor drag on both trajectories independently of which trajectory the rotor-drag coefficients were estimated on.
This confirms that our approach is applicable to any feasible trajectory once the drag coefficients are identified, which is an advantage over methods that improve tracking performance for a specific trajectory only (e.g. \cite{Hehn14mech}).
With the rotor-drag coefficients estimated on the circle trajectory, we achieve almost the same performance in terms of absolute trajectory tracking error on the lemniscate trajectory as with the coefficients identified on the lemniscate trajectory but not vice versa.
As discussed above, this is due to a higher excitation in body velocities on the circle trajectory which results in a better identification of the rotor-drag coefficients.
This suggests to perform the rotor-drag coefficients identification on a trajectory that maximally excites the body velocities.

\begin{table}[b]
\caption{Maximum and standard deviation of the position error $\poserr$ as well as the absolute trajectory tracking error $\abserr$~\eqref{eq:absolute_error} over ten loops on both the circle and the lemniscate trajectory.
For each trajectory, we perform the experiment without considering drag, with the drag coefficients identified on the circle trajectory, and with the drag coefficients identified on the lemniscate trajectory.}
\label{table:trajectory_tracking_statistics}
	\centering
	\begin{tabular}{|c|c|ccc|}
		\multirow{2}{*}{Trajectory} & \multirow{2}{*}{Params ID} & $\max \left( \norm{\poserr} \right)$ & $\sigma \left( \norm{\poserr} \right)$ & \abserr \\ 
		& & [\SI{}{\centi\meter}] & [\SI{}{\centi\meter}] & [\SI{}{\centi\meter}] \\ \hline
		\multirow{3}{*}{Circle} & Not Cons. Drag & 21.08 & \textbf{2.11} & 17.53 \\ \cline{2-5}
		& Circle & 14.54 & 2.63 & ~\textbf{6.54} \\ \cline{2-5}
		& Lemniscate & \textbf{12.39} & 2.53 & ~8.16 \\ \hline \hline
		\multirow{3}{*}{Lemniscate} & Not Cons. Drag & 16.79 & 3.19 & 11.27 \\ \cline{2-5}
		& Circle & 10.25 & 2.30 & ~5.56 \\ \cline{2-5}
		& Lemniscate & \textbf{10.02} & \textbf{2.23} & ~\textbf{5.51} \\ \hline
	\end{tabular}
\end{table}

\begin{figure}[]
   \centering
   \includegraphics[width=0.48\textwidth]{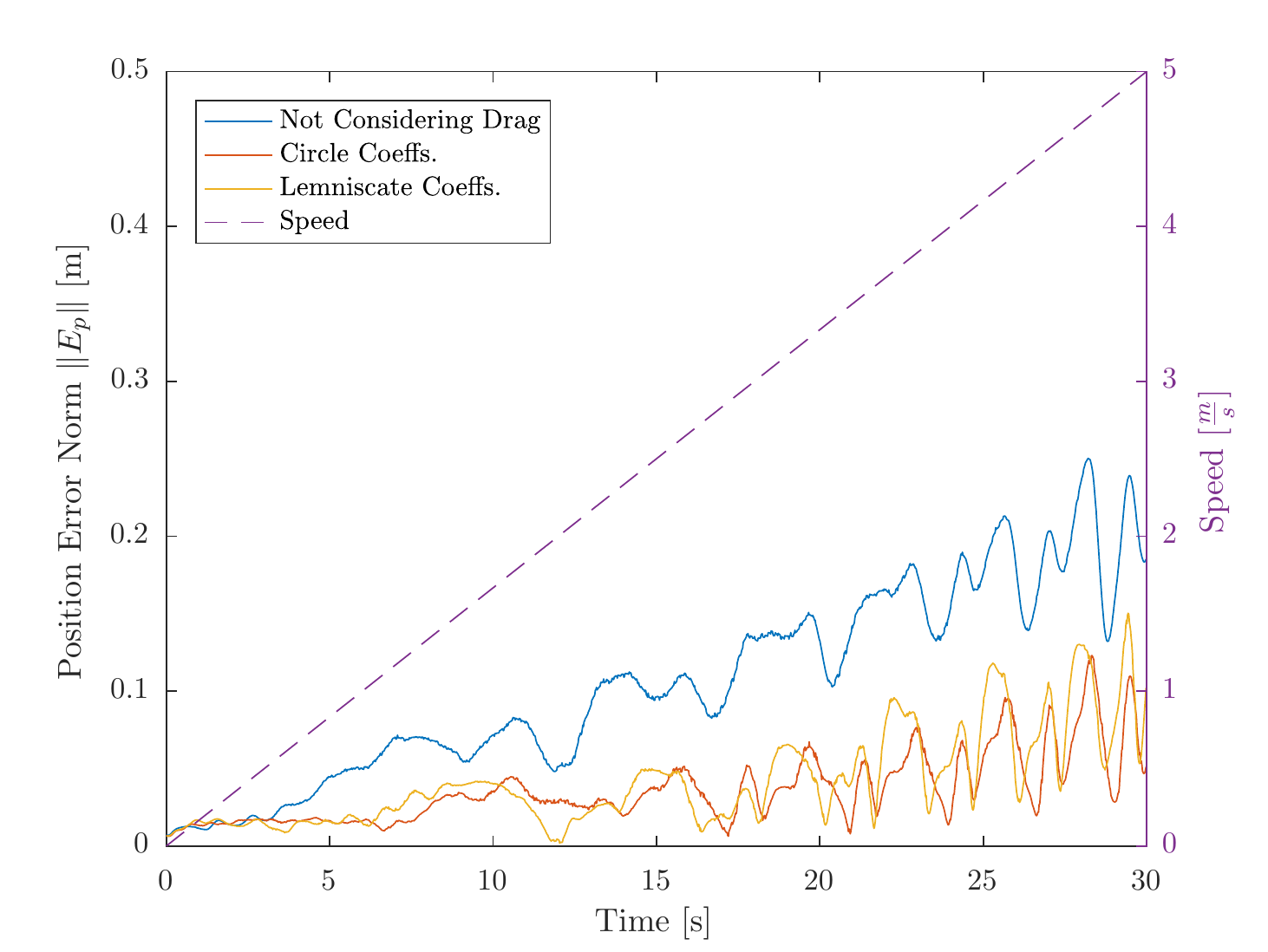}
   \caption{Position error norm $\norm{\poserr}$ when ramping the speed on the circle trajectory from \SI{0}{\meter\per\second} up to \SI{5}{\meter\per\second} in \SI{30}{\second} without considering rotor drag (solid blue), with drag coefficients estimated on the circle trajectory (solid red), and with drag coefficients estimated on the lemniscate trajectory (solid yellow).
   The reference speed on the trajectory is shown in dashed purple.}
   \label{fig:circle_ramp_up}
\end{figure}

\begin{figure}[]
   \centering
   \includegraphics[width=0.48\textwidth]{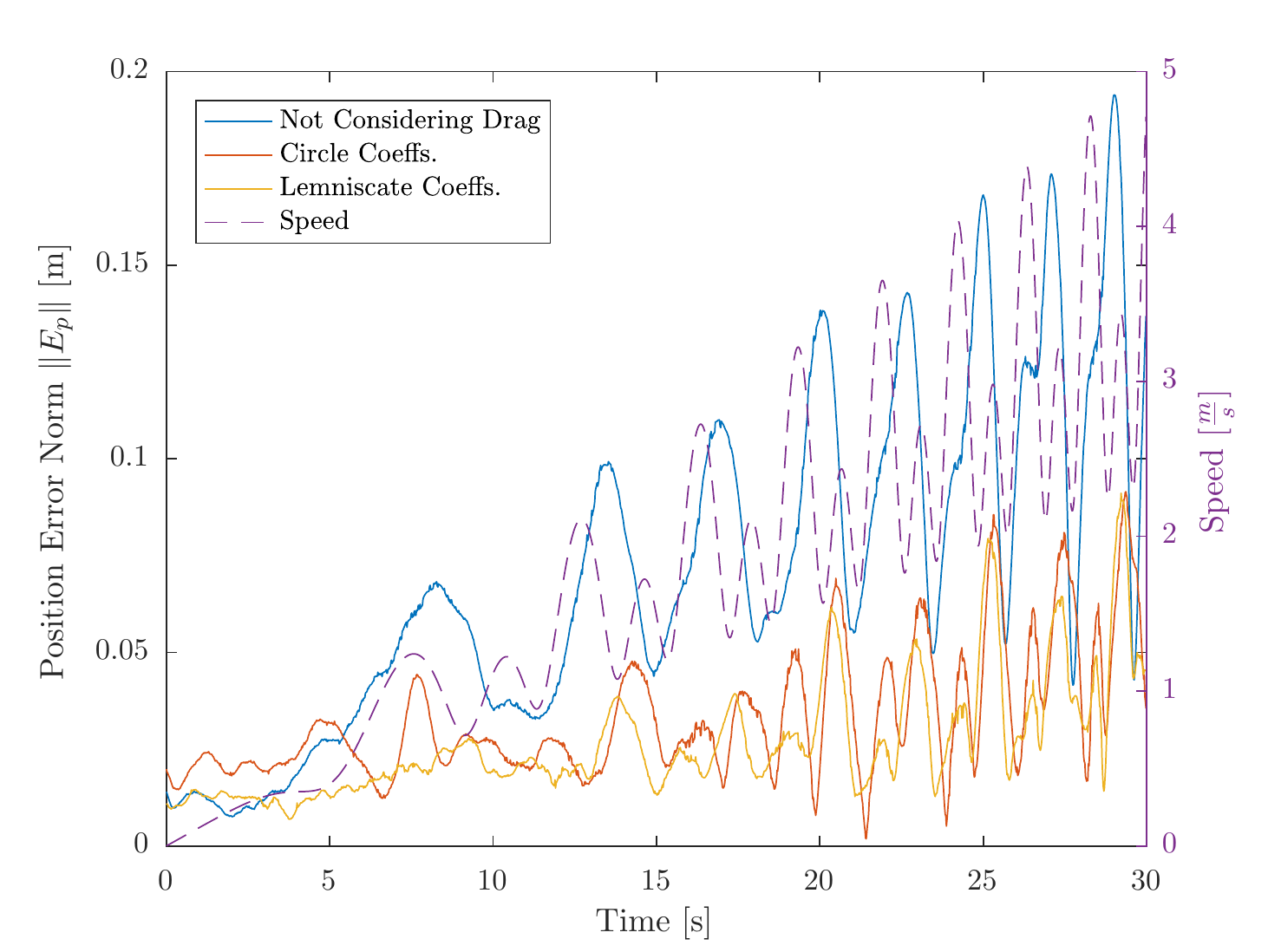}
   \caption{Position error norm $\norm{\poserr}$ when ramping the maximum speed on the lemniscate trajectory from \SI{0}{\meter\per\second} up to \SI{5}{\meter\per\second} in \SI{30}{\second} without considering rotor drag (solid blue), with drag coefficients estimated on the circle trajectory (solid red), and with drag coefficients estimated on the lemniscate trajectory (solid yellow).
   The reference speed on the trajectory is shown in dashed purple.}
   \label{fig:figure_eight_ramp_up}
\end{figure}

Since the rotor drag is a function of the velocity of the quadrotor, we show the benefits of our control approach by linearly ramping up the maximum speed on both trajectories from \SI{0}{\meter\per\second} to \SI{5}{\meter\per\second} in \SI{30}{\second}.
Fig.~\ref{fig:circle_ramp_up}, and Fig.~\ref{fig:figure_eight_ramp_up} show the position error norm and the reference speed over time until the desired maximum speed of \SI{5}{\meter\per\second} is reached for the circle and the lemniscate trajectory, respectively.
Both figures show that considering rotor drag does not improve trajectory tracking for small speeds below \SI{0.5}{\meter\per\second} but noticeably does so for higher speeds.

An analysis of the remaining position error for the case where rotor drag is considered reveals that it strongly correlates to the applied collective thrust.
In the used dynamical model of a quadrotor, we assume the rotor drag to be independent of the thrust [c.f.~\eqref{eq:velocity_dynamics}], which is not true in reality.
For the experiment in Fig.~\ref{fig:circle_ramp_up}, the commanded mass-normalized collective thrust varies between \SI{10}{\meter\per\second\squared} and \SI{18}{\meter\per\second\squared} which clearly violates the constant thrust assumption.
By considering a dependency of the rotor drag on the thrust might improve trajectory tracking even further and is subject of future work.

%% file: chapters/control_comparison.tex
\section{Comparison to Other Control Methods} \label{sec:control_comparison}

In this section, we present a qualitative comparison to other quadrotor controllers that consider rotor drag effects as presented in~\cite{Kai17ifac}, \cite{Omari13iros}, \cite{Svacha17icuas}, and \cite{Bangura17thesis}.

None of these works show or exploit the differential flatness property of quadrotor dynamics subject to rotor drag effects.
They also do not consider asymmetric vehicles where ${\dragcoeff{x} \neq \dragcoeff{y}}$ and they omit the computation of $\bodyrate_z$ and $\dot{\bodyrate}_z$.

In~\cite{Kai17ifac} and~\cite{Omari13iros}, the presented position controller decomposes the rotor drag force into a component that is independent of the vehicle's orientation and one along the thrust direction, which leads to an explicit expression for the desired thrust direction.
They both neglect feed-forward on angular accelerations, which does not allow perfect trajectory tracking.
As in our work, \cite{Kai17ifac} models the rotor drag to be proportional to the square root of the thrust, which is proportional to the rotor speed, but then assumes the thrust to be constant for the computation of the rotor drag, whereas~\cite{Omari13iros} models the rotor drag to be proportional to the thrust.
Simulation results are presented in~\cite{Kai17ifac} while real experiments with speeds of up to \SI{2.5}{\meter\per\second} were conducted in~\cite{Omari13iros}.

The controller in~\cite{Svacha17icuas} considers rotor drag in the computation of the thrust command and the desired orientation but it does not use feed-forward terms on body rates and angular accelerations, which does not allow perfect trajectory tracking.
In our work, we use the same thrust model but neglect its dependency on the rotor speed.
In~\cite{Svacha17icuas}, also the rotor drag is modeled to depend on the rotor speed, which is physically correct but requires the rotor speeds to be measured for it to be considered in the controller.
They present real experiments with speeds of up to \SI{4.0}{\meter\per\second} and unlike us also show trajectory tracking improvements in vertical flight.

As in our work, \cite{Bangura17thesis} considers rotor drag for the computation of the desired thrust, orientation, body rates, and angular accelerations.
However, their computations rely on a model where the rotor drag is proportional to the rotor thrust.
Also, they neglect the snap of the trajectory and instead require the estimated acceleration and jerk, which are typically not available, for computing the desired body rates and angular accelerations.
The presented results in~\cite{Bangura17thesis} stem from real experiments with speeds of up to \SI{1.0}{\meter\per\second}.

%% file: chapters/conclusion.tex
\section{Conclusion} \label{sec:conclusion}

We proved that the dynamical model of a quadrotor subject to linear rotor drag effects is differentially flat.
This property was exploited to compute feed-forward control terms as algebraic functions of a reference trajectory to be tracked.
We presented a control policy that uses these feed-forward terms, which compensates for rotor drag effects, and therefore improves the trajectory tracking performance of a quadrotor already from speeds of \SI{0.5}{\meter\per\second} onwards.
The proposed control method reduces the root mean squared tracking error by \SI{50}{\percent} independently of the executed trajectory, which we showed by evaluating the tracking performance of a circle and a lemniscate trajectory.
In future work, we want to consider the dependency of the rotor drag on the applied thrust to further improve the trajectory tracking performance of quadrotors.